# Combining the Strengths of Dutch Survey and Register Data in a Data Challenge to Predict Fertility (PreFer)

*Unpublished manuscript*

Elizaveta Sivak[1,2,*], Paulina Pankowska[3], Adriënne Mendrik[4], Tom Emery[5], Javier Garcia-Bernardo[6], Seyit Höcük[7], Kasia Karpinska[5], Angelica Maineri[5], Joris Mulder[7], Malvina Nissim[8], Gert Stulp[1,2]

[1] Department of Sociology, University of Groningen, Grote Rozenstraat 31, 9712TS, Groningen, the Netherlands

[2] Inter-University Center for Social Science Theory and Methodology, University of Groningen, Grote Rozenstraat 31, 9712TS, Groningen, the Netherlands

[3] Department of Sociology, Utrecht University, Padualaan 14, 3584CH, Utrecht, The Netherlands

[4] Eyra, Saturnusstraat 14 - Unit 4.13, 2516AH, The Hague, The Netherlands

[5] Erasmus School of Social and Behavioral Sciences, Erasmus University Rotterdam, Thomas Morelaan, 3062PA, Rotterdam, The Netherlands

[6] Department of Methodology and Statistics, Utrecht University, Padualaan 14, 3584CH, Utrecht, The Netherlands

[7] Centerdata, Tilburg University, Warandelaan 2, 5037AB, Tilburg, The Netherlands

[8] Center for Language and Cognition Groningen, Faculty of Arts, University of Groningen, Oude Kijk in 't Jatstraat 26, 9712EK, Groningen, The Netherlands

* Author for correspondence, e.sivak@rug.nl

## Abstract

The social sciences have produced an impressive body of research on determinants of fertility outcomes, or whether and when people have children. However, the strength of these determinants and underlying theories are rarely evaluated on their predictive ability on new data. This prevents us from systematically comparing studies, hindering the evaluation and accumulation of knowledge. In this paper, we present two datasets which can be used to study the predictability of fertility outcomes in the Netherlands. One dataset is based on the LISS panel, a longitudinal survey which includes thousands of variables on a wide range of topics, including individual preferences and values. The other is based on the Dutch register data which lacks attitudinal data but includes detailed information about the life courses of millions of Dutch residents. We provide information about the datasets and the samples, and describe the fertility outcome of interest. We also introduce the fertility prediction data challenge PreFer which is based on these datasets and will start in Spring 2024. We outline the ways in which measuring the predictability of fertility outcomes using these datasets and combining their strengths in the data challenge can advance our understanding of fertility behaviour and computational social science. We further provide details for participants on how to take part in the data challenge.

**Keywords**: fertility, data challenge, benchmark, out-of-sample prediction, survey data, register data

# 1 Introduction

Fertility outcomes – or whether and when people have children – is a major topic of study across the human sciences because of its importance for individuals and societies. Sociological and demographic research has developed numerous theories of fertility [1–6] and produced a sophisticated body of work on the many characteristics associated with fertility [7]. These range from the social environment during upbringing [8,9] to partnership trajectories in later life [10,11], from social interactions with friends [12,13] to family policies in society [14], and from biological differences [15–17] to differences in values [5,18]. Despite these advancements in our understanding of fertility outcomes, there is no agreement on the relative importance of these characteristics [19–23]. Moreover, characteristics considered important only explain a fraction of the variation in fertility outcomes [24,25], and factors thought to underlie fertility declines cannot explain recent drops in fertility [26]. This suggests that our understanding of fertility is still limited.

In the social sciences, there is a growing recognition that quantifying (out-of-sample) predictability of an outcome can improve our scientific understanding of it and assess the practical relevance of the theories explaining it [27–31]. Despite the potential of a focus on prediction, it remains under-utilised in the social sciences and demography in particular, although notable exceptions do exist [32–38]. One of the methods to measure predictability is a data challenge, where several teams compete to predict a particular outcome using the same dataset and evaluation criteria. Data challenges have led to major progress in different disciplines [39–41], but rarely have been used in the social sciences.

In this paper, we present two unique data sources which can be used to measure the predictability of fertility outcomes to help overcome some of the problems of fertility research and describe how to use these datasets in a data challenge. One of these data sources is the LISS panel, a longitudinal survey based on a random, representative sample of the Dutch population covering a wide range of topics, including many factors associated with fertility, identified in the previous research. The other is Dutch register data, which includes information about the life courses of the entire Dutch population. Using this combination of "wide" survey data and "long" administrative data within a data challenge framework can provide insights for fertility research, social policy, and family planning. Furthermore, it may help to clarify the reasons behind poor predictions of other life outcomes [35,42], and it can serve as a showcase for how data science methods can advance our understanding of different phenomena of interest to the social sciences.

The main aims of the paper are the following. First, to describe these datasets and the data preprocessing steps that we took tailored for the task of measuring predictability of fertility outcomes. Second, to introduce the data challenge PreFer for predicting fertility outcomes in the Netherlands which uses these datasets. We outline the potential benefits of the data challenge in understanding fertility behaviour, present its methodology, and provide details on how to participate in the challenge. Before we do so, we first describe the advantages of the focus on prediction and using data challenges in the social sciences.

### 1.1 Explanatory and predictive modelling

A dominant approach in the social sciences, including fertility research, is explanatory modelling. A typical statistical model assesses a pre-specified theoretical model on the basis of a limited number of variables, and support for a theoretical mechanism is often based on whether an estimated coefficient is different from zero, most often assessed via a p-value. The quality of the model is traditionally evaluated on the same data that was used to estimate the model.

While this approach has advanced and will continue to advance our scientific understanding, it also has shortcomings. First, the process of evaluating the quality of the model using the same data the model was fitted on may result in overfitting, in which a model may pick up on peculiarities in the data that do not generalise to other, unseen cases [28,43]. This means that we likely put too much confidence in the findings arising from a model that is fitted and evaluated on the same data. Second, the inclusion of a limited number of variables and in addition a limited number of interactions between them (often for the sake of interpretability) may mean important variables and non-linear relationships are overlooked (i.e., underfitting) [29], and the importance of different factors cannot be assessed. Third, while the p-value, when statistical assumptions are met, can show whether an estimate is unlikely to be zero, it cannot serve as a measure of effect size [27], and thus cannot be used to determine which variables are most strongly associated with an outcome across different models (this also holds for frequently used effect sizes like the odds-ratio [44,45]). The p-value is also easily influenced by decisions in the process of statistical analysis, including sample selection, outlier removal, or the operationalisation of variables [31,46].

Given these limitations of the p-value, underfitting, and overfitting, it is harder to systematically compare different studies and assess the practical importance of specific theories for social policy. These limitations are also partially responsible for the reproducibility crisis observed in many disciplines [47–51].

Complementing explanatory modelling with predictive modelling – using a statistical model to predict previously unseen observations and measure the predictive accuracy – may alleviate these problems [27]. Out-of-sample predictive ability, or how well a model can predict novel cases (e.g., out-of-sample root mean squared error, out-of-sample accuracy in predicting binary outcomes), is an easy-to-understand and useful measure of model quality. It has the same interpretation, regardless of the underlying assumptions of the statistical model. For example, the predictive ability of a Poisson regression, linear regression, or decision tree on the same outcome using the same predictor variables can be usefully compared. Out-of-sample predictions help avoid overfitting and hence false positive results because it is evaluated on novel data (on a held-out set or using cross-validation, a process in which a dataset is separated into several training and test sets and the quality of the model is determined by its performance across the different test sets). All that makes out-of-sample predictive ability a better measure of how well our model is performing and to what extent our theories are predictive in the real world, producing more valid evidence for practical use [52–54].

A data-driven approach focusing on prediction can further alleviate underfitting, because such an approach often includes many or all variables available (of course, at a risk of hindering interpretability and causal analysis). This also allows us to assess how much each predictor contributes to the model predictions, compare the importance of a wider set of predictors, and find novel predictors. Many data-driven analytical approaches are also well-suited to identify non-linear patterns and interactions [29]. Overfitting of complex data-driven models can again be guarded against through a process of cross-validation. In the case of overfitting, the model can be simplified (for example using regularization). Variability in results across different datasets in the process of cross-validation (e.g., which variables are selected in the model in different iterations) by itself indicates which variables are consistently associated with the outcome of interest.

**1.2 Data challenges**

One of the ways in which a focus on predictive ability has led to rapid progress in other disciplines is through data challenges [55,56], also known as benchmarks or common tasks. A data challenge consists of inviting (teams of) researchers to engage in a common task of trying to best predict a particular outcome in the holdout dataset on the basis of a common training dataset using a pre-defined metric for out-of-sample predictive ability [55].

Data challenges have led to advancements and breakthroughs in several scientific fields, including computer and data science [40], natural language processing [41,57,58], physics [59], biology [60], and biomedicine [39]. Such challenges allow us to assess the limits of predictability of an outcome given the data and methods (i.e., statistical analysis strategies). When many researchers with various backgrounds participate in a data challenge, the final result likely reflects not just the limits of a particular method or skills of researchers, but the current limits of predictability for a given dataset [35]. The element of competition (driven by the publication of ranking, desire to beat the current high score, public acknowledgement of winning teams, and, sometimes, prize money) and getting access to normally restricted datasets motivate people to participate and publicly evaluate methods in terms of predictive performance (sometimes referred to as "benchmarking"), which aids in better estimating the upper limits of predictability.

Data challenges can also accelerate scientific progress because they allow us to compare different methods, and through this comparison gain insights into the research problem at hand [35,56]. For example, gaps in predictive ability between theory-driven models (based on smaller sets of variables specified in theories) and data-driven models can prompt discussions as to why these gaps exist and stimulate improvements in theories, data, and measurements [35,42,53]. A comparison of models can identify predictive yet overlooked variables, best operationalisations of variables, non-linear effects, and interactions between variables.

In the social sciences to date, one large-scale data challenge has been organised, namely the Fragile Families Challenge [35]. In this challenge, participants predicted six life outcomes of adolescents in the United States, using a longitudinal dataset with thousands of early life predictors from birth cohort surveys. The challenge showed low predictability of these life outcomes: the best predictive models were only slightly better than simply predicting the mean of the training data. One of the major conclusions of this landmark study was that our understanding of these life outcomes may have been more limited than previously thought. This sparked discussions on the reasons for low predictability [42], ranging from acknowledging that previous scientific understanding of child development is incomplete or incorrect to the hypothesis that these outcomes are inherently unpredictable to the idea that the sample size typical for social science surveys was not sufficiently large for machine learning algorithms to produce accurate predictions [35,42].

**1.3 Novel opportunities for fertility research**

Using a combination of the LISS panel data and Dutch register data in a data challenge framework can significantly impact fertility research for several reasons. The first benefit comes from measuring the current limits of predictability of fertility outcomes. This is an end in itself, as we currently do not know how predictive common variables in fertility research are. The (in-sample) measures of model quality that are occasionally presented (e.g. the coefficient of determination) can be a poor proxy of the strength of out-of-sample predictive ability, and this holds even more for p-values that are often used as evidence for the strength of a particular theoretical mechanism [27]. Measures of out-of-sample predictability also constitute a better basis for cumulative scientific progress [28]. In future analyses based on the same datasets, the predictive ability of novel methods (e.g., selection of variables, improved algorithms) can be compared to established benchmarks based on the challenge.

A combination of "wide" survey data (many variables/features) and "long" administrative data (many cases) provides a good opportunity to measure the current predictability of fertility outcomes. The LISS panel includes many of the previously identified factors associated with fertility behaviour, including intentions and values. Dutch register data includes many important variables (measured over twenty years) for the entire population. Recent developments in register data have furthermore led to the opportunity of creating many additional variables. For example, variables can be created on the basis of information on neighbourhood characteristics, characteristics of the workplace, and, rather uniquely, on people's social networks (e.g., through information on neighbours, kin, colleagues, and classmates). The setup of our data challenge further allows linking the survey data to the register data and combining their strengths to increase predictive ability (see Combining Survey and Register Data).

The second benefit is that a quantification of the predictive ability of various variables helps determine the scope of potential interventions. A highly significant variable that has low predictive value is not a useful target for intervention. Identifying the most important predictors helps create a shortlist of potential interventions that can then be tested independently. This can also help individuals make more informed decisions concerning family planning and avoid having fewer children than desired, which is common in Western countries [61,62].

Further opportunities for fertility research come from comparing and interpreting different methods employed within the data challenge. In particular, comparing data-driven methods to theory-driven can contribute to theorising in several ways. The differences in predictive ability between theory- and data-driven methods highlight possible improvements for theorising based on the dataset and variables at hand [53]. Such improvements can come from overlooked variables, non-linear effects, and interactions between variables that are less systematically evaluated in theory-driven analyses. Comparing the predictive performance and most predictive variables for the survey and register data gives insight into the importance of different types of data, e.g. detailed longitudinal data about life courses spanning about two decades or rich data about attitudes, preferences, and values.

An advantage of predictive modelling is that it readily allows for assessing for which groups of people predictions are best (or worst). Such post-hoc predictive performance analyses can provide new insights into the reasons for varying performance [63]. The large sample size of the Dutch register data allows such detailed analysis. These analyses are made stronger through a data challenge because it can give insights into whether the behaviour of some groups is predicted well and other groups poorly by all analytical strategies, or whether analytical strategies vary in which groups they can predict well.

## 2 Data description

**2.1 LISS panel survey data**

The LISS panel is a high-quality online survey infrastructure based on a traditional probability sample drawn from the Dutch population register by Statistics Netherlands and is managed by the non-profit research institute Centerdata. The representativeness of the LISS panel is similar to that of traditional surveys based on probability sampling[1] [64,65]. Initial selection biases were substantially corrected by refreshment samples [66].

There are two main sources of data on the LISS panel: the LISS Core Study and Background surveys. The LISS Core Study is a longitudinal study that is fielded each year in the LISS panel and measures the same set of variables. The Core Study includes ten modules that cover a wide range of topics from income, education, and health to values, religion, and personality, including variables designed specifically to study fertility behaviour (e.g. fertility intentions)[2]. The Background survey is filled out by a household's contact person when the household joins the panel and is updated monthly[3]. It collects basic socio-demographic information about the household and all of its members (including those who are not LISS panel members and do not participate in the Core surveys). The description of the LISS Core Study modules and Background survey is provided in Table 1.

---

[1] Details about the sample, recruitment, and refreshment samples can be found at https://www.lissdata.nl/methodology.

The Core Study modules and all their different waves are stored separately. For the task of measuring the predictability of fertility outcomes, we constructed a merged dataset based on all modules from the LISS Core Study from 2007-2020. This dataset consists of more than 30 thousand variables.

The task of the data challenge is to predict who will have a child in 2021-2023 based on data from all previous years (see Fertility outcome and Methodology for details). As very few people have children before the age of 18 and after the age of 45, we chose as the target group those who were between 18 and 45 years old in 2020 and who participated in at least one Core study in 2007-2020.

The LISS panel started in 2007 when approximately 5000 households comprising 8000 individuals of 16 years and older were recruited (about 6000 of them being 18-45 years old) [67]. The annual attrition rate is approximately 10%. To counteract this drop out, new panel members are recruited every two years based on the population registers (i.e., refreshment samples), maintaining the representativeness of the LISS panel [66]. Overall, in 2007-2020 around ten thousand people aged 18-45 were members (at least at some point) of the households recruited in the LISS panel. When members from recruited households moved out of the household they remained in the panel (but in a different household). About 70% of this group (~ 6900 people) actually participated in at least one Core survey between 2007 and 2020. These people are our target group, and all of them are included in our main dataset.

Most of our target group, or LISS panel members who participated in at least one Core study before 2020 and were aged 18-45 in 2020, have dropped out of the LISS panel by 2021-2023. To create our outcome variable, we could make use of both the Core surveys and the Background variables, but even still we were able to create the outcome for only about 1400 respondents (99% of these respondents participated at least in one Core study in 2019-2020, so for almost all of them the most recent predictors are available). For a data challenge, this number is rather small; however, it is a common sample size for a social science dataset on a representative sample. Moreover, there are no alternative options for survey data that have longitudinally gathered so much data from respondents and that can be linked to register data.

---

[2] The questionnaires of all the Core Study modules can be found at https://www.dataarchive.lissdata.nl/study-units/view/1.
[3] The questionnaire of the Background survey is available at https://doi.org/10.57990/qn3k-as78.

The dataset is split into a *training set* (the outcome and predictors), available to participants of the challenge (around 70% of the whole dataset) and a *holdout set* for evaluation (the remaining 30% of the dataset), unavailable to the participants during the data challenge (see Fig. 1).

An important consideration in creating training and holdout data is how to deal with participants from the same household. Participants from the same household cannot be considered independent data points. Using models fitted on particular people in the household in the training data to make predictions about other people in the household in the holdout data can be seen as a case of overfitting, as there are several variables measured on the household level that have identical values for all household members, and a model can pick up on these similarities to make predictions. This is why we randomly selected households rather than participants into the training or holdout data meaning that all participants of one household are either in the training data or in the holdout data.

To do that we selected the households where the outcome was available at least for one household member and grouped these households into two groups: 1) where at least one person had a new child, 2) where no one had a new child. Then we randomly selected 30% households from each group. We assigned all participants who belong to these households to the holdout set, and excluded people for whom the outcome is missing from the holdout set. All participants from the remaining 70% of households (as well as participants from the households where the outcome was missing) were assigned to the training set. To verify whether the participants in the resulting training and holdout groups are similar, we compared the distributions of three variables in the holdout and training sets (excluding participants with a missing outcome): the outcome, age, and the number of waves of the Core Study modules a person participated in (operationalised as answering at least one question). Participants in the training and holdout data were very similar based on these variables. It is important to note that only part of the training set—people for whom the outcome is not missing—is comparable to the holdout set because the holdout set does not include people for whom the outcome is missing and the outcome is probably not missing at random.

We also provide two additional datasets which optionally can be used by the data challenge participants to enrich the training set (see Fig. 1). The first is the Background variables dataset, which includes all monthly values of several variables from the Background survey for the duration that a respondent (or household) participated in the panel. This dataset includes information from all members of the LISS panel (including those who are not in our target group) and their household members. The second additional dataset is based on the Core Study modules and is in its structure identical to the main dataset but contains information on respondents who are younger than 18 and older than 45.

Participants will also have access to two machine-readable codebooks that contain information on, amongst other things, how particular variables have been measured over time, possible answer options to each question, and the type of variables (e.g., categorical, numerical, or date). These codebooks have been created specifically for the PreFer data challenge. Current codebooks for the LISS panel are separate for each survey and in either pdf-format or on a password protected website, which can limit the efficiency of data preparation for most machine learning approaches [68].

**Table 1** LISS panel studies that are included in the merged dataset. The codebooks (in Dutch and English) are available via the links

| Name | Description | DOI |
|---|---|---|
| Background variables | Socio-demographic variables at the household level and individual level. Filled in by a contact person about all the household members participating in the LISS panel when the household joins the panel. Thereafter, the contact person is presented with the background questionnaire every month to enter any changes that may have occurred. | https://doi.org/10.57990/qn3k-as78 |
| Core Study modules: | | |
|    Health | Physical and mental health assessments and medication use, lifestyle habits. | https://doi.org/10.17026/dans-ze3-5uk9 |
|    Religion and Ethnicity | Religious upbringing, religious affiliation, religiosity, religious orthodoxy. Nationality, origin, ethnic identification, language proficiency and use. | https://doi.org/10.17026/dans-xkw-t8dm |
|    Social Integration and Leisure | Social contacts, core discussion network, loneliness. Leisure activities, voluntary work and informal care, social media usage. | https://doi.org/10.17026/dans-zaf-casa |
|    Family and Household | Family structure, social support from family, parenting and children, domestic responsibilities, child education and childcare. | https://doi.org/10.17026/dans-xkd-5hp5 |
|    Work and Schooling | Employment status and history, job satisfaction and conditions. Education, qualifications, and training. | https://doi.org/10.17026/dans-x26-tttv |
|    Personality | Subjective well-being, personality traits. | https://doi.org/10.17026/dans-x5h-4cxd |
|    Politics and Values | Political Engagement and Attitudes, Political Affiliation and Orientation, Values and Social Attitudes. | https://doi.org/10.17026/dans-zms-r5rz |
|    Economic Situation: Assets | Different kinds of assets, loans and debts. | https://doi.org/10.17026/dans-z2r-n69z |
|    Economic Situation: Income | Different sources of income, subjective standard of living. | https://doi.org/10.17026/dans-24y-dkqk |

| Economic Situation: Housing | Housing characteristics, expenditures, satisfaction with housing. | https://doi.org/10.17026/dans-zgv-9qky |

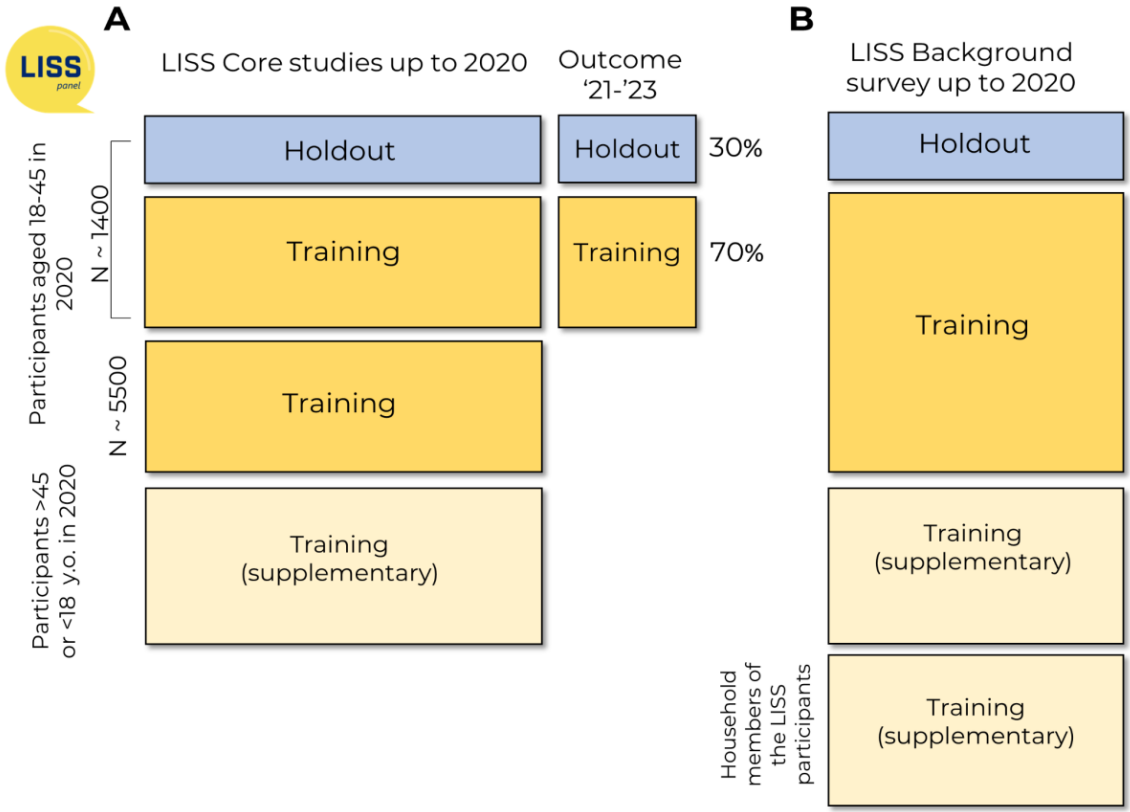

**Fig. 1** Survey data from the LISS panel used in the data challenge. **A)** Two datasets based on the Core Study modules from 2007-2020. The main dataset contains only of the target group: participants of the LISS panel aged 18-45 in 2020, for whom at least some information is available in these Core study modules (~6800 people). The outcome is available for ~1400 of them. A supplementary dataset containing the same Core Study modules but only for respondents who are younger than 18 and older than 45 is provided in a separate file. **C)** Background dataset which contains monthly information on about 30 variables from the LISS Background survey from 2007-2020 for all LISS panel participants and their household members.

**2.2 Dutch register data (CBS)**

The register data comes from several Dutch registers collected by Statistics Netherlands (CBS) (we will refer to this source as CBS data) [69]. It includes many datasets about persons, households, jobs, businesses, dwellings, vehicles, and more[4].

For the task of measuring the predictability of fertility outcomes, we selected and merged several CBS datasets. We did not make use of the datasets that appear less relevant (e.g., about businesses) and those that contain particularly sensitive information (e.g., prescribed medication). The list of datasets that are available during the challenge with a brief description is provided in Table 2. Most of these datasets cover the period from 1995 to 2023. These datasets include information about marriages and partnerships, children, education, employment, income and assets, neighbourhood characteristics and more. A dataset is also available on 1.4 billion relationships between all 17 million inhabitants of the Netherlands [70], leading to a unique opportunity to include information on how people are embedded in networks of family members, neighbours, colleagues, household members, and classmates, and on characteristics of people in these networks.

Based on these selected datasets, we prepared a starter package: a base preprocessed dataset (mostly with the data from 2020) along with a codebook in Dutch and English. This dataset contains information about all individuals who were: 1) 18-45 years old at the end of 2020 (because of the outcome we have chosen, see Fertility Outcome), and 2) residents of the Netherlands at least in 2020-2023 (i.e., for whom we can establish the fertility outcome and for whom at least some information from previous years is available). In addition to the variables already included in the selected datasets (such as level of education, partnership status, and personal income), we constructed more than twenty variables for this sample (e.g., age, total number of children in 2020, age of the youngest child in 2020, total number of marriages and partnerships by the end of 2020, characteristics of jobs). Moreover, for each individual in this dataset, we added information on the household level (e.g., household income and composition), on the partner if the focal individual had one (e.g., partner's education, income and socio-economic category), and on neighbourhood characteristics (e.g., distance to the closest childcare). We also linked results of the Dutch 2017 general elections, 2019 provincial elections, and 2020 municipal elections (proportion of votes for different parties by municipality) as voting for particular parties might correlate with conservative views and religion [71,72].

Participants of the data challenge will be able to calculate additional variables based on the full longitudinal datasets that are available (see Table 2). As an example, the training data may be enhanced by characteristics of the networks of the participant, using the network datasets for linkage [70]. Example scripts will be available on the challenge website[5] on how to preprocess network datasets and calculate network characteristics.

---

[4] For the full list of datasets available, see the CBS micro-data catalogue (https://www.cbs.nl/nl-nl/onze-diensten/maatwerk-en-microdata/microdata-zelf-onderzoek-doen/catalogus-microdata).
[5] http://preferdatachallenge.nl.

Additional CBS datasets (not initially selected) can be requested throughout the challenge with a short justification of why the dataset is requested. The relevant CBS datasets can be searched using the CBS micro-data catalogue[6] and ODISSEI portal[7]. Data from external sources (not included in the CBS datasets) that can be linked to groups of individuals can also be uploaded (if approved by trained CBS employees) – for example, welfare policies by municipality[8].

We split the sample into training (70%) and holdout data (30%). We first randomly split the households, meaning that individuals within one household are all either in the training data or the holdout data. Then we randomly split the holdout set into the data for the intermediate leaderboard (one third of the holdout set; 10% of the entire sample) and the data for the final leaderboard (two thirds; 20% of the entire sample). All intermediate submissions will be assessed on the intermediate leaderboard set, and only the predictive performance of the final submissions will be assessed on the final leaderboard set. The size of the CBS dataset allows setting aside this intermediate leaderboard set to allow more submissions before the final one without the increased risk of overfitting.

To allow adding the characteristics of the networks of individuals in the dataset and because of the submission process (see Submission), only the outcome from the holdout data is withheld; other variables will be available for the whole sample and also for people over 45 and under 18 years of age (see Fig. 2).

It should be noted that CBS has not been involved in the design of this study and access to the CBS data within the data challenge is subject to clearance of CBS.

**Table 2** The list of CBS datasets that will be available to the participants of the data challenge. The full codebooks (in pdf format, in Dutch) are available via the links to the CBS website. Brief descriptions and lists of variables are also available via the DOI links to the ODISSEI portal

| Name of the dataset | Description | DOI |
|---|---|---|
| Gbapersoontab [CBS] | Personal characteristics of people in the population registry (BRP) | https://doi.org/10.57934/0b01e4108071ba40 |
| Gbaburgerlijkestaat Bus [CBS] | Civil status of persons included in the BRP (marriages, registered partnerships) | https://doi.org/10.57934/0b01e410803b37e0 |

---

[6]The catalogue is available at https://www.cbs.nl/nl-nl/onze-diensten/maatwerk-en-microdata/microdata-zelf-onderzoek-doen/catalogus-microdata.
[7] See the instructions on how to use the ODISSEI portal (https://portal.odissei.nl/) in the user guide https://guides.dataverse.org/en/5.13/user/.
[8] See the instructions for uploading external datasets here https://www.cbs.nl/en-gb/our-services/customised-services-microdata/microdata-conducting-your-own-research/importing-external-datasets.

| | | |
|---|---|---|
| Gbaverbintenispartnerbus [CBS] | Partner's IDs to link partner's characteristics | https://doi.org/10.57934/0b01e410801f93bf |
| Gbamigratiegebeurtenisbus [CBS] | Migration dates | https://doi.org/10.57934/0b01e4108021261a |
| Gbahuishoudensbus [CBS] | Household id and characteristics | https://doi.org/10.57934/0b01e410802125bb |
| Hoogsteopltab [CBS] | Highest achieved (meaning: with diploma) and followed (meaning: without a diploma) level of education | https://doi.org/10.57934/0b01e410801fd716 |
| Inpatab [CBS] | Personal income | https://doi.org/10.57934/0b01e41080372fbd |
| Inhatab [CBS] | Household income | https://doi.org/10.57934/0b01e41080371196 |
| Vehtab [CBS] | Household wealth | https://doi.org/10.57934/0b01e4108037363f |
| Koppelpersoonhuishouden | To link households' income and wealth to individuals | DOI not available yet |
| Spolisbus [CBS] | Employment (excluding self-employed) | https://doi.org/10.57934/0b01e410804cb681 |
| Secmbus [CBS] | Personal socio-economic category (employed, self-employed, studying, etc.) | https://doi.org/10.57934/0b01e410803432a6 |
| Nabijheidkindopvtab [CBS] | Proximity to childcare (from objects, e.g. living places) | https://doi.org/10.57934/0b01e41080238887 |
| Gbaadresobjectbus [CBS] | Objects numbers and personal IDs (to link objects' characteristics to individuals) | https://doi.org/10.57934/0b01e410802154d6 |
| Vbowoningtypetab [CBS] | Type of housing | https://doi.org/10.57934/0b01e4108053c864 |
| Vslgwbtab [CBS] | Municipality and neighborhood codes of residence objects to link external data about municipalities and neighborhoods. | https://doi.org/10.57934/0b01e41080236a82 |
| **Networks:** | | |
| Burennetwerktab [CBS] | Neighbors network | DOI not available yet |
| Colleganetwerktab [CBS] | Colleagues network | DOI not available yet |
| Familienetwerktab [CBS] | Family network | DOI not available yet |
| Huisgenotennetwerktab [CBS] | Household network | DOI not available yet |
| Klasgenotennetwerktab [CBS] | Classmates network | DOI not available yet |
| **External data**: results of Dutch election | Results of the Dutch general elections (2017), provincial elections (2019, and municipal elections (2020) by municipality | DOI not available |

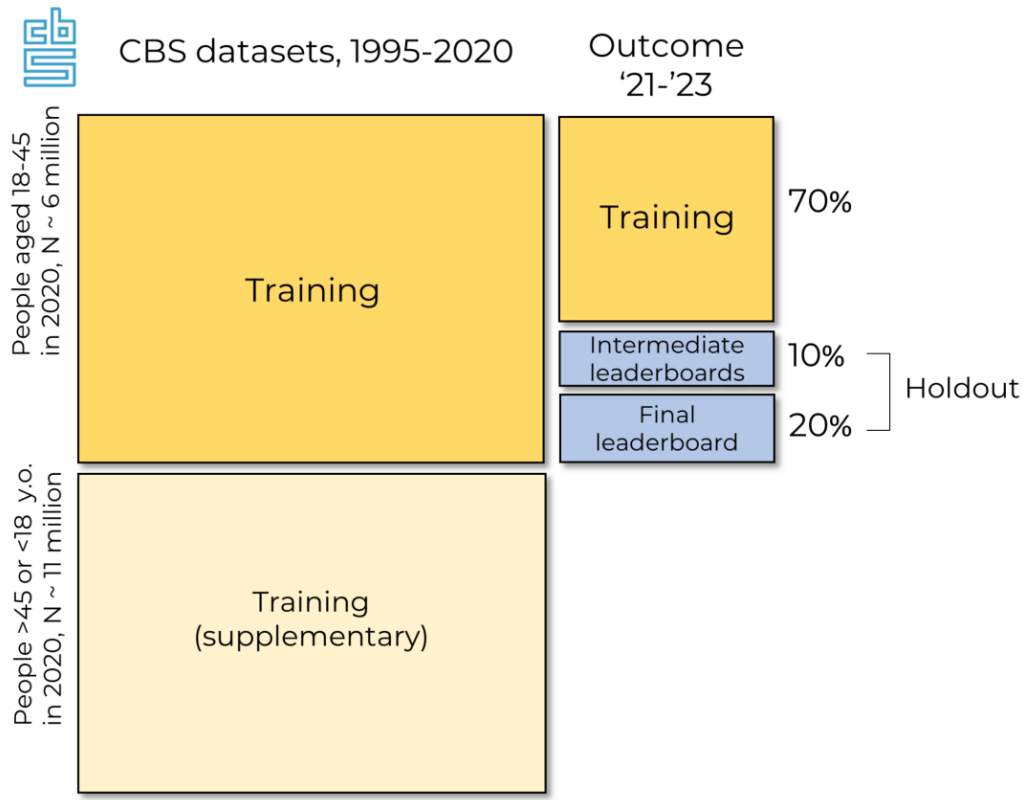

**Fig. 2** The scheme of the CBS data used in the challenge. The target group includes Dutch residents aged 18-45 in 2020. For them, part of the outcome variable (70%) and background variables are available for training. The background data is also available for the older and younger age groups which may be needed to calculate particular network characteristics of the people in the target group

## 2.3 Combining survey and register data

The LISS data can be linked to CBS data inside the secure Remote Access (RA) CBS environment. Almost all LISS participants consented to this linkage. We performed the linking and this was successful for approximately 90% of LISS panel participants. This linkage provides a unique opportunity to develop and test multiple approaches to enhance the predictive performance by using both datasets. For example, the LISS training data can be enriched by adding variables about the families of the panel respondents, information which is available inside the CBS RA. Moreover, missing values in the outcome can be imputed from CBS data to increase the LISS training set. Transfer learning [73] can also be used to leverage the strength of both datasets; this would involve first training using the register data with subsequent fine-tuning on survey data. Such approaches possibly yield better predictions on the LISS holdout set.

## 2.4 Fertility outcome

For the data challenge, in both datasets we constructed the following fertility outcome: having a(nother) child in 2021-2023, either biological or adopted.

We chose this outcome for several reasons. First, it is a hard task, but the shorter time frame makes it less difficult than predicting fertility outcomes that unfold over a much longer period, such as age at first birth or the number of children. The studies on the association between intentions to have a child in the future – argued to be strong determinants of reproductive behaviour [1] – and actual fertility illustrate the difficulties in predicting long-term outcomes. There is a well-established discrepancy between lifetime fertility intentions, or the total intended family size, and completed fertility outcomes in low-fertility settings [74–76]. Changing life circumstances, macrostructural shocks, and uncertainty and instability of fertility intentions themselves over the life course likely account for this discrepancy [77–83]. A shorter time frame reduces the chance of these changes, making short-term fertility intentions more likely to be realised (and short-term fertility potentially more predictable), although the degree of their realisation varies by country [75,84–89].

Second, different processes may underlie births of different parity (e.g., first child, second child) [90], which the data challenge can tap into [91]. For example, the fertility behaviour of siblings particularly strongly affected respondents' first but not second births [9]. In contrast, closer spatial proximity to kin increases the likelihood of second births and decreases first births [92].

A third and pragmatic reason is that of data availability. Several fertility outcomes that are also of interest, such as the age at first or last birth or the total number of children can only be derived for the population that has already reached the end of their reproductive period (i.e., at least 45 years old, born in 1975 or earlier). The LISS panel started in 2007 when people born in those cohorts were already 32 or older. Potential important information about individuals' life courses is either unavailable or available in retrospect and may therefore not be reliable [93–95]. This also holds true to some extent for the Dutch register data, as many important variables such as education are only available from 1995-1999 onwards, so for the cohorts born before 1975-1979 this data is scarce or unavailable. For this reason, attempting to predict whether respondents have a child in a longer subsequent period (e.g., 10 years), would also come at a cost of data availability, as substantial proportions of LISS respondents will not have data available for over ten years.

A final reason for choosing this particular outcome is the potential practical utility. The postponement of childbirth is a major cause of involuntary childlessness [96] and the increased demand for medically assisted reproduction. An increased understanding and better prediction of rates in which couples are not able to realise their fertility intentions can be used in quantifying future need for assisted reproductive technologies.

To create the outcome for the LISS data, we primarily used the data from the "Family and household" Core Study module. We used information on the number of children in 2020-2023 (alive and deceased) and the relation between the parents and children (i.e., biological, adoptive, step-parent or foster parent). We used information from the background variables dataset if information on the outcome was missing based on the "Family and household" module. On the basis of these variables, we calculated a binary outcome: whether a person had at least one new child in 2021-2023 or not. Parts of the LISS data (including the "Family and Household" 2023 wave) will be made openly available for researchers only after the end of the data challenge.

In the case of the CBS data, we used the CBS dataset Kindoudertab[9] which links children with their legal parents. Based on that, for each person in the sample (Dutch residents aged 18-45 in 2020), we calculated the number of children in each year between 2021 and 2023 and then derived whether or not a person had at least one new child in 2021-2023.

Approximately 25% of people in the LISS dataset (for whom the outcome is known) and approximately 15%[10] in the CBS dataset had a new child between 2021 and 2023. The percentage for the LISS panel is higher because of the way we constructed the outcome for this dataset. With particular patterns of missing and available data, we can be certain whether respondents had a child but we cannot be certain about them not having a child. For instance, if information for a respondent is missing in 2023, we cannot exclude the possibility that this respondent had a child in 2023. Conversely, any increase in the number of children in 2021, 2022, or 2023 means that a new child was born even if information for some of the waves is missing. Therefore, among individuals with incomplete data on the number of children, we can only determine the outcome for some who had a new child (during the years for which information is available), leaving those without a new child underrepresented among the part of our LISS sample for whom the outcome is known.

# 3 Methodology of the data challenge PreFer

Here we describe the Predicting Fertility (PreFer) data challenge. For the most recent updates and further details, see the PreFer website https://preferdatachallenge.nl.

## 3.1 The task, goals and research questions

The goal of the data challenge is to assess the current predictability of individual-level fertility and improve our understanding of fertility behaviour.

This challenge focuses on the following task: predict for people aged 18-45 in 2020, who will have a(nother) child within the following three years (2021-2023) based on the data up to and including 2020.

---

[9] Details about the dataset can be found at https://doi.org/10.57934/0b01e410801f9401.
[10] At the time of writing the most recent data from 2023 on parent-child links has not yet been released so this number is an approximation.

The results of the data challenge will be used to answer the following research questions:

- How well can we predict who will have a(nother) child in the short-term future in the Netherlands?
- What are the most important predictors of this fertility outcome?
- Are there novel predictors for this fertility outcome, unaccounted for in the existing theoretical literature? (this can include non-linear effects and interactions between predictors)
- How do theory-driven methods compare to data-driven methods in terms of predictive accuracy?
- What poses larger constraints on predictive ability: the number of cases or the number of ('subjective') variables? Survey data typically consists of hundreds or thousands of variables (including subjective measures like intentions or values) on a relatively small sample (at least in comparison to data science projects [42]). Population registries typically contain fewer variables only on a set of 'objective' measures (e.g., income, education, cohabitation) but describe a large number of people.
- To what extent can predictions on survey data be improved by augmenting it by register data? (e.g. imputing missing values, correcting measurement errors, adding new variables)
- To what extent can predictions based on the register data be improved by augmenting it with survey data (e.g. "subjective" variables)?

**3.2 Phases of the challenge**

The challenge includes two phases (Fig. 3). The first phase is predicting the outcome using only the LISS data. Participants will be able to download the LISS training data on their own devices and run their methods locally. They will submit their methods through a submission platform (see Submission). The first phase will take place in April-May 2024.

In the beginning of June 2024, Phase 2, which includes three tracks, will start. Based on the results of the first phase, several of the best-performing teams will be selected for tracks 1 and 2 of the second phase to work inside the secure Remote Access (RA) CBS environment. The second phase will run until the end of September 2024. Teams that are not selected into tracks 1 and 2 will continue working on the LISS data (this is track 3).

Access to the CBS RA environment and CBS data is governed by strict rules and regulations in relation to data protection and privacy. One consequence of such rules is that access to this RA environment is only possible from the European Economic Area and a few other countries[11] and is subject to the approval of CBS and passing security checks. Another issue in working in the CBS RA environment is that computing resources are constrained. Given the limitations, only a selection of teams can participate in the second phase. Around 10-20 teams will be selected from the first phase into tracks 1 and 2 of the second phase and will be allowed to access the CBS RA environment (see Determining the Winners for how the teams will be selected). The costs of access to the CBS datasets will be covered by ODISSEI and access will be subject to the vetting and agreement of Statistics Netherlands and the ODISSEI Management Board under the general grant conditions of ODISSEI.

---

[11] See the full list of countries at https://commission.europa.eu/law/law-topic/data-protection/international-dimension-data-protection/adequacy-decisions_en.

Tracks 1 and 2 differ on the holdout set for which the participants will predict the outcome. Participants themselves can choose which track(s) they will work on. In the first track, participants will predict the fertility outcome for the LISS holdout set. This is similar to Phase 1/track 3, but the difference is that the LISS data can be linked to CBS data inside the RA environment. In the second track, participants will instead predict the fertility outcome for the CBS holdout set. This setup provides the participants of tracks 1 and 2 with a unique opportunity to develop and test multiple approaches to possibly enhance the performance of their methods by using both datasets (see Combining Survey and Register Data).

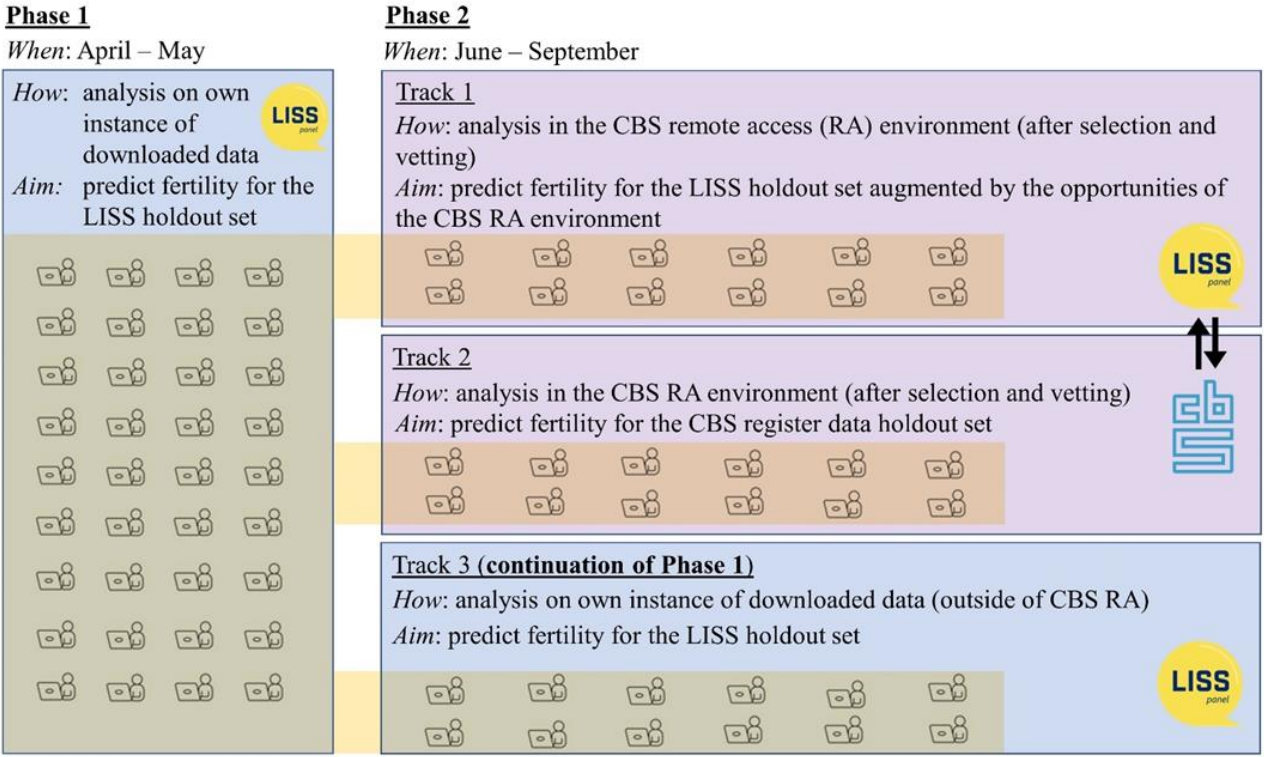

**Fig. 3** Phases of the PreFer data challenge

### 3.3 Submission

In the first phase and in track 3 in Phase 2, participants are asked to submit their methods (the trained model and code that needs to be applied to the holdout data, as well as the code used for training) rather than the predicted values themselves, along with a description of the method used (e.g. approach to selecting the variables and machine learning model and preprocessing the data). If participants performed analyses to interrogate their model (see Determining the Winners), for example, assessing the importance of different predictors for different groups, these scripts should be provided as well.

For the submissions, participants will use the open-source web-platform Next. It allows for reproducible submissions in data challenges in which data is not publicly available, and therefore common solutions like Kaggle are not possible. Instructions on how to submit to the platform and example code will be provided on the PreFer website preferdatachallenge.nl. The submission platform supports the programming languages Python and R. Potential submissions are automatically run on example data to check for errors. If these checks are successful then the method can be submitted and will be evaluated on the holdout dataset. This workflow fosters computational reproducibility, which was a concern in the Fragile Families Challenge in which participants only submitted their predictions [97]. This also allows us to run submitted methods on different (or future) variants of the data.

In tracks 1 and 2 of the second phase, participants cannot make use of the submission platform because the register data is only available within the CBS Remote Access environment. The participants are asked to submit predicted values generated by their method by saving them in a specified folder inside the RA environment along with the trained model, all scripts used for data preprocessing and model training, and a description of the method.

All the methods submitted in the challenge will be made publicly available in a GitHub repository[12] as well as the PreFer website. The models based on the CBS dataset will only be made public after they have been screened by CBS to ensure that the code itself does not disclose identifiable information.

**3.4 Evaluation**

3.4.1 Metrics

The metrics below are used in both phases of the challenge to assess the quality of the predictions (i.e., the difference between the predicted values and the ground truth). These are common metrics for classification tasks (i.e., predicting binary outcomes).

**Accuracy:** The ratio of correct predictions to the total number of predictions made.
 Accuracy = # correct predictions / total # predictions

---

[12] https://github.com/eyra/fertility-prediction-challenge.

**Precision:** The proportion of positive predictions that were actually correct (i.e., the proportion of people who actually had a new child in 2021-2023 of all the people who were predicted to have a new child in this period).

Precision = # true positives / (# true positives + # false positives)

**Recall:** The proportion of positive cases that were correctly identified (i.e., the proportion of people who actually had a new child and were predicted to have a new child of all people in the sample who had a new child in 2021-2023).

Recall = # true positives / (# true positives + # false negatives)

**F1 score (for the positive class, or having a new child):** The harmonic mean of the precision (P) and recall (R).

F1 = 2∗(Precision∗Recall) / (Precision+Recall)

For both phases of the data challenge, all four metrics will be used for the leaderboards (ranked lists of the predictive performance of the submitted methods on the holdout data). The F1 score leaderboard is the main leaderboard that will be used as the quantitative criteria to determine the winners of the challenge. We chose the F1 score as the main metric because we are interested in methods that achieve an overall good performance in distinguishing between those who had and did not have a child in 2021-2023. The F1 score helps develop methods which strike a good balance between recall and precision, or that are reliable in identifying people who had a new child and at the same time tries to minimize the number of false positive predictions. This will allow us to better understand what predicts having and not having a child. Accuracy is less suitable in this case because of the class imbalance, i.e., the relatively low proportion of those who had a new child (around 25% in the LISS data, 15% in the CBS data).

To prevent overfitting to the holdout data, the number of submissions during the challenge will be limited. Before the final submissions, participants will be able to make several intermediate submissions in each phase (the number of submissions and the deadlines will be provided on the data challenge website), after each of them the in-between, anonymous leaderboards will be presented.

### 3.4.2 Determining the winners

To achieve the goals of the challenge, the winners are determined using both quantitative and qualitative criteria. For the research goal of determining the predictability of the fertility outcome, we use a quantitative evaluation, as described above. That is, for each of the three tracks (track 1, track 2, and phase 1 together with track 3), a winner will be determined on the basis of the F1 score. Overall, there will be three winners determined based on the F1 score.

The F1 score will also be used as the main selection criterion for entry into tracks 1 and 2 of the second phase, for which approximately 10-20 teams will be selected. However, the LISS and CBS datasets may require different skills to achieve the best result. For example, some algorithms might perform worse on the LISS dataset but might benefit from the larger sample size of the CBS dataset in terms of performance. To ensure the representation of different methods in the second phase, an evaluation committee will assess the submissions with the top F1 scores to select the teams that can proceed to tracks 1 and 2 (provided that at least one team member can be present for at least a part of Phase 2 in a country where it is allowed to access the CBS RA and this person also passes security checks and is approved by CBS). The evaluation committee will consist of the organisers of the challenge, an expert in fertility research, and a data scientist.

To recognise other important contributions in furthering the understanding of fertility behaviour, an evaluation committee will also assess the submissions on the basis of qualitative criteria: 1) innovativeness: a novel approach using ideas from either social sciences or data science (e.g. using approaches such as transfer learning, still uncommon in the social sciences), and 2) whether the method improves our understanding of fertility. The latter can be done by unpacking the method, for example, by doing error analysis, or examining misclassified cases and trying to understand why the method failed to classify them; analysing predictive performance for particular groups; analysing interactions and importance of factors overall and for different groups; identifying good predictors that were not considered so far. Overall, two additional winners (one for each criterion) among all challenge participants will be selected based on these qualitative criteria.

All winners (five teams) will have an opportunity to present their method and results in a plenary session at the ODISSEI Conference for Social Science in the Netherlands in Autumn 2024. One representative per team will have the costs of attending the conference covered.

It is important to note that while we will select winners to recognise particular contributions and to encourage the development of the best possible methods during the data challenge, the goals of the data challenge can only be achieved through community efforts of all the participants of the challenge. Because of that all the submissions are highly valued and will be recognised in scientific publications based on the challenge (see A Special Issue).

## 3.5 Ethics

Predicting individual life outcomes can be a sensitive topic. However, we believe that the potential benefits of this data challenge outweigh the potential risks. The main potential benefit is more robust knowledge about one of the most important life outcomes that is at the heart of many governmental policies [98]. Importantly, a substantial part of the group that is studied in the data challenge (people aged 18-45 living in the Netherlands) can benefit from the challenge, for example, by learning more about the key factors that can hinder them from achieving their desired family size. In particular, involuntary childlessness can have serious consequences for well-being.

The data challenge itself does not appear to substantially increase the risk of privacy breaches, because all data used in the challenge is either already available (or will be available soon after the challenge) in the case of the LISS panel or access to it is very strictly managed in the case of CBS[13]. Nonetheless, for the LISS panel, the risk of de-anonymization may be increased. First, the over 120 datasets that previously needed to be separately downloaded and linked will now be presented as one merged file to participants. Second, the advertisement of the challenge may reach people who would otherwise not have engaged with the LISS panel data.

To evaluate and combat risks of identification, a data protection impact assessment (DPIA) was carried out for the LISS dataset by Centerdata, the institution responsible for the management of the longitudinal survey. A DPIA is a structured procedure to identify potential risks at an early stage associated with the handling of personal data. It serves as a crucial instrument for risk mitigation and for showcasing adherence to GDPR compliance standards. The potential risks of privacy breaches, the likelihood of their occurrence, and the potential impact they would have were identified so that appropriate additional measures could be taken to mitigate these risks. Subsequently, the levels of residual risks (the remaining risk after appropriate measures have been taken) were assessed, revealing no medium or high-level residual risks.

The measures already implemented by Centerdata, following its standard procedures for disseminating survey data to the LISS Data Archive (e.g., pseudonymization, data cleaning, data aggregation, and exclusion of sensitive personal data in open answers), already adhere to the GDPR requirements and comply with Centerdata's privacy policy[14]. As an additional measure, the datasets used for the data challenge were further pseudonymized with a unique respondent ID specific to this project. This means that participants of the data challenge cannot link the data used in the challenge to additional data in the LISS Data Archive. Moreover, although participation in the challenge is open to anyone who wants to participate, pre-registration with a name and email address is necessary. The data used for the data challenge are stored in a secure and closed environment on the Next platform. Registered participants will be invited to read and agree to a LISS data user statement specifically tailored for this project, describing what is permitted and prohibited when working with the data. Only after agreeing to these terms and conditions are they allowed to download the data for the challenge.

With respect to the CBS data, the data will only be available for a small group of participants within the Remote Access environment, conditional on passing a security and awareness test, where all exports are verified and strict rules regulate what can and cannot be exported from the environment[15]. All access will be managed through the standard CBS access protocols with each researcher being evaluated individually and all current safeguards maintained. All directly identifying personal details are removed from the CBS datasets and replaced by a pseudo key. There are also additional precautions in place, such as data minimisation (e.g., exact date of birth and income information not made available). Furthermore, to prevent de-anonymisation, the CBS data cannot be enriched with other data, unless this linkage with external data is approved by trained CBS employees.

---

[13] The measures to protect personal data and the data privacy regulations that CBS adheres to are described here: https://www.cbs.nl/en-gb/about-us/who-we-are/our-organisation/privacy.

[14] The privacy statement can be found at https://www.centerdata.nl/en/privacy-statement.

[15] See the rules concerning the export of information from the CBS RA environment here: https://www.cbs.nl/en-gb/our-services/customised-services-microdata/microdata-conducting-your-own-research/export-of-information.

Another potential risk concerns the misuse of the predictive methods developed in the data challenge that can pose threats to one's privacy, especially if the accuracy of predictions is high. For example, businesses may be interested to know when employees or customers are likely to have children as in the infamous case where a retail customer's pregnancy was predicted based on previous consumption behaviour and baby products were directed at the customer [99]. We believe these risks are mitigated by the fact that if predictive accuracy is high, it will likely require data on many variables. Such extensive data at the individual level is difficult to acquire outside a research setting, and cannot be collected without a person's knowledge and consent.

### 3.6 Feasibility assessment and constraints

To test the setup and infrastructure of the data challenge we organised a pilot data challenge at the Summer School for Computational Social Science at ODISSEI in 2023 (SICSS-ODISSEI). The methodology was similar to one of the upcoming data challenge. The teams used two datasets (first LISS then CBS[16]) to predict having a(nother) child within the next three years (2020-2022) based on data up to and including 2019.

A first version of the infrastructure was tested and lessons learned are taken into account for the subsequent version of the infrastructure that will be used during the upcoming data challenge. Overall, the infrastructure worked well, but some more detailed participant instructions for the submission process will be added for Phase 1 of the upcoming challenge. Furthermore, the participants will be provided with updated documentation of the LISS and CBS datasets and instructions on how to work with several CBS datasets such as social network files. Based on the participant's experiences, we made a FAQ about the submission process and using the CBS RA as well as a list of common problems during the submission process and how to deal with them, both of which will be posted on the PreFer website.

Some drawbacks of our setup are harder to overcome. For example, while much effort has been put into allowing researchers access to CBS data, there are limitations in terms of the programming languages and versions that it can provide, the descriptions of the different datasets that can in principle be used, and the computing resources (limited storage and memory and slower computations in peak hours). On the website, we further describe these constraints and how participants can deal with them.

### 3.7 About the organisers and participation in the data challenge

The PreFer data challenge is organised by a collaboration between the Department of Sociology at the University of Groningen; ODISSEI, the national research infrastructure for the social sciences in the Netherlands; Eyra, a developer of software-as-a-service solutions for reproducible science; and Centerdata, a research institute managing the LISS panel[17]. The team includes academic researchers, data scientists, survey methodologists, and software engineers.

---

[16] Project number 9469.

## 4 A special issue with the results of the data challenge

We plan to publish a paper presenting the design and results of the PreFer data challenge. Everyone who was part of a team that made a working submission at least in one phase of the challenge will be invited to be a co-author of this paper. By a working submission we mean a submitted method that produced predictions for the holdout set and that is accompanied by a description of the method. There will be no limit on the number of participants who can qualify as co-authors.

Additionally, we plan to publish a special issue on the results of the data challenge. All the participants of the data challenge will be invited to submit a manuscript to this special issue. The submitted papers will be peer-reviewed.

The call for papers with detailed instructions and requirements will be published later on the PreFer website. A paper should describe the process that led to the final submission. This includes for example decisions concerning data preprocessing and handling missing data, model and variable selection, and what was learned during this process. A paper can also be aimed at describing how the data challenge contributed to fertility research. Other ideas will also be possible after discussing them with the challenge organisers. Manuscripts need to be accompanied by a clearly documented modular open-source code that will allow other researchers to reproduce all the results, as well as figures and tables in the article.

## 5 Data availability

**Access for the participants of the PreFer data challenge**

During the challenge, all PreFer participants will be able to download the LISS training dataset, the background variables dataset, and the dataset with information from individuals not included in the target group via a link provided after registration and after signing a data user statement. Access to the CBS data is only granted after a vetting procedure (see Phases of the Challenge).

**Access outside of the PreFer data challenge**

Most LISS panel data (except the 2023 wave of the Family and Household survey needed to calculate the outcome variable for the data challenge and recent Background information) can already be accessed for non-commercial scientific or policy-relevant purposes by researchers affiliated with academic institutions after signing a data user statement. Data are deliberately withheld until after the data challenge.

---

[17] Further details about the organizers can be found at http://preferdatachallenge.nl.

The scripts used to create all the LISS training datasets and the holdout dataset (including the script to calculate the outcome variable) will be available on the project page in the LISS data archive[18] approximately in October 2024, after PreFer ends.

Researchers affiliated with a number of authorised scientific organisations can get access to the CBS data for scientific purposes[19]. The code to produce the outcome variable, reproduce the train-test split, and prepare the base dataset will also be available at the same page in the LISS data archive[20].

# 6 Acknowledgements

We are thankful to the participants of the pilot data challenge at SICSS-ODISSEI 2023 for their feedback that helped to improve the data challenge. We also thank Priscilla Zhang and Mara Verheijen from Centerdata for merging and preparing the raw data files from the longitudinal LISS Core Study and the LISS background surveys.

# 7 Funding

This work is supported by a VIDI grant (VI.Vidi.201.119) from the Netherlands Organization for Scientific Research (NWO) to GS. The LISS panel data were collected by the non-profit research institute Centerdata (Tilburg University, the Netherlands). Funding for the panel's ongoing operations comes from the Domain Plan SSH and ODISSEI since 2019. The initial set-up of the LISS panel in 2007 was funded through the MESS project by the Netherlands Organization for Scientific Research (NWO). The ODISSEI Benchmark Platform, the ODISSEI-SICSS Summer School, and the development of the LISS harmonized dataset are financed by the ODISSEI Roadmap Project financed by NWO.

---

[18] This is the link to the project page in the LISS data archive: https://doi.org/10.57990/f3ge-3a61.
[19] The list of the authorised institutions can be found at https://www.cbs.nl/en-gb/our-services/customised-services-microdata/microdata-conducting-your-own-research/institutions-and-projects. The application process is described here: https://www.cbs.nl/en-gb/our-services/customised-services-microdata/microdata-conducting-your-own-research/applying-for-access-to-microdata.
[20] See the project page in the LISS Data Archive here: https://doi.org/10.57990/f3ge-3a61.